\documentclass[conference]{IEEEtran}

\usepackage{graphicx}
\usepackage{amsmath,amssymb,amsfonts}
\usepackage{booktabs}
\usepackage{times}
\usepackage{microtype}
\usepackage{bm}

\usepackage[hidelinks]{hyperref}

\title{\Large\bfseries
White-Box Neural Ensemble for Vehicular Plasticity: Quantifying the Efficiency Cost of Symbolic Auditability in Adaptive NMPC
}

\author{
  \IEEEauthorblockN{Enzo Nicolás Spotorno,
  Matheus Wagner, and
  Antônio Augusto Fröhlich}
  \IEEEauthorblockA{\textit{Federal University of Santa Catarina (UFSC)}\\
  \textit{Florianópolis, Brazil}\\
  Emails:\texttt{\{enzoniko, wagner, guto\}@lisha.ufsc.br}
  }
}

\let\svthefootnote\thefootnote
\newcommand\freefootnote[1]{%
  \let\thefootnote\relax%
  \footnotetext{#1}%
  \let\thefootnote\svthefootnote%
}

\begin{document}
\maketitle

\begin{abstract}
We present a white-box adaptive NMPC architecture that resolves vehicular plasticity (adaptation to varying operating regimes without retraining) by arbitrating among frozen, regime-specific neural specialists using a \textit{Modular Sovereignty} paradigm. The ensemble dynamics are maintained as a fully traversable symbolic graph in CasADi, enabling maximal runtime auditability. Synchronous simulation validates rapid adaptation ($\sim$7.3\,ms) and near-ideal tracking fidelity under compound regime shifts (friction, mass, drag) where non-adaptive baselines fail. Empirical benchmarking quantifies the transparency cost: symbolic graph maintenance increases solver latency by 72--102$\times$ versus compiled parametric physics models, establishing the efficiency price of strict white-box implementation. \freefootnote{This work was partially supported by FUNDEP grants Rota 2030 27192.02.INT01/2022.01-00, 29271.02.01/2022.01-00 and 29271.03.01/2023.04-00.}
\end{abstract}

\begin{IEEEkeywords} Adaptive Model Predictive Control, Neural Networks, Autonomous Vehicles, Symbolic Auditability, Vehicular Plasticity.
\end{IEEEkeywords}

\section{Introduction and Related Work}
\label{sec:intro}

Autonomous systems, particularly in safety-critical domains such as automotive stability control and autonomous racing, face persistent challenges related to the plasticity and verifiability of the dynamic models employed across their control and decision-making components. On one hand, such systems must exhibit plasticity without retraining, adapting their behavior to variations in operating conditions and fleet heterogeneity to maintain high performance \cite{maiworm2021online}. On the other hand, certification assurance processes under standards such as ISO 26262 favor transparent, analyzable white-box models \cite{salay2018analysis}. State-of-the-art deployment tools (e.g., \textit{acados}, L4CasADi \cite{salzmann2024learning}) achieve real-time performance by emitting compiled model code, effectively avoiding runtime symbolic traversal and shifting verification to code-generation provenance and reproducible-build practices \cite{verschueren2022acados}, satisfying ISO 26262 via tool qualification. In contrast, we investigate the alternative of retaining an explicit symbolic representation (e.g., via CasADi SX \cite{andersson2019casadi}) at runtime. This approach offers a maximal theoretical bound on transparency, allowing direct inspection of the derivative graph, but incurs a computational overhead that remains under-characterized.

Recent learning-based modeling approaches, influenced by trends toward Universal Foundation Models \cite{spotorno2026positioncertifiablestateintegrity}, often rely on fine-tuning large, monolithic models. Such approaches are susceptible to the stability–plasticity dilemma, potentially compromising closed-loop stability and formal guarantees \cite{maheshwari2025addressingplasticitystabilitydilemmareinforcement, wu2023composing}. In contrast, learning-based approaches such as Mixture-of-Experts models and the HYDRA architecture \cite{spotorno2026positioncertifiablestateintegrity}, which serves as the inspiration for the \textit{Modular Sovereignty} paradigm adopted in this work, advocate systems that adapt by arbitrating among a library of frozen, pre-verified specialist models, rather than continuously updating a single complex black box \cite{pan2025controllable}. This modular structure preserves the integrity of individual components and remains compatible with symbolic implementations, thereby supporting both adaptation and verification.

Related work, such as implicit adaptation methods (e.g., Neural-Fly \cite{o2022neural} and PDE-informed function-approximation approaches like S$^2$GPT-PINN \cite{ji2025s}) adjust behavior via expressive function approximators, while explicit parameter adaptation approaches (e.g., Deep Dynamics \cite{chrosniak2024deep}) estimate physical parameters within a fixed structure. Real-time approaches include LPV-MPC \cite{morato2023predictive}, GP-MPC \cite{2502.02310}, and hierarchical adaptive NMPC \cite{2304.12263} (demonstrated at $\sim$35\,ms solve time on full-scale vehicles). Although monolithic foundation models may offer broad generalization, safety-critical applications often favor modularity to facilitate formal verification and to mitigate catastrophic forgetting \cite{spotorno2026positioncertifiablestateintegrity}. Moreover, monolithic networks exhibit spectral bias, preferentially fitting low-frequency components first, which can obscure abrupt regime changes important for control \cite{rahaman2019spectral}. Our approach resembles Mixture-of-Experts architectures but replaces opaque, learned gating with an optimization-based \textit{Governor} QP that blends frozen, regime-specific experts. This prioritizes regime separation and makes abrupt changes analyzable rather than implicitly smoothed \cite{spotorno2026positioncertifiablestateintegrity}.

Nevertheless, a gap remains in understanding the practical implications of this paradigm, particularly the performance costs incurred when enforcing symbolic implementations within NMPC solvers. Distinct from efficiency-optimized methods, our work isolates and quantifies the specific efficiency cost of maintaining full non-linear symbolic retention within a \textit{Modular Sovereignty} architecture \cite{spotorno2026positioncertifiablestateintegrity}. We aim to assess the trade-offs associated with integrating such a neural adaptive controller into a white-box NMPC framework, prioritizing symbolic inspectability, traceability, and full auditability over raw execution speed. Our contributions are threefold: we present a symbolic NMPC integration of a Sparse ensemble of frozen Neural Specialists designed for vehicular plasticity under strict auditability constraints; we characterize the computational cost of this symbolic transparency, quantifying the efficiency gap relative to parametric models; and we functionally validate the architecture's ability to handle abrupt regime shifts (e.g., friction changes) and fleet heterogeneity. We demonstrate via performance bound analysis that the method resolves the plasticity problem despite the identified computational cost.
\section{Methodology}

We adopt a HYDRA-inspired \textit{Modular Sovereignty} architecture \cite{spotorno2026positioncertifiablestateintegrity}. Rather than training a monolithic model or explicitly estimating physical parameters, we construct the vehicle dynamics as a convex combination of frozen, regime-specific specialist vector fields, each parameterizing the state derivative of a continuous-time dynamics model. Global effective dynamics use vector field mixing: $\dot{\hat{\mathbf{x}}}(t) = \sum_{i=1}^{N} w_i \cdot \Psi_i(\mathbf{x}(t), \mathbf{u}(t))$, where $\mathbf{w} \in \mathbb{R}^N$ is the governance vector ($\sum_i w_i = 1, w_i \ge 0$), $\Psi_i$ are frozen specialist networks, and $\mathbf{x}, \mathbf{u}$ are state and input. This constrains dynamics to the specialists' pointwise convex hull, avoiding extrapolation beyond the model family. This geometric property is central to handling fleet heterogeneity: by spanning the space of dynamics between disparate platforms, the convex combination allows the controller to interpolate the correct dynamics for an unseen vehicle configuration without retraining. The adaptive NMPC shares the kinematic states (position and orientation) while the dynamic accelerations (longitudinal, lateral, and yaw-rate derivatives) are computed as a convex combination of specialist outputs, where weights are estimated via the \textit{Governor} \cite{spotorno2026positioncertifiablestateintegrity} using finite-difference approximations of measured state derivatives ($\dot{\bm{x}} \approx (\bm{x}_k - \bm{x}_{k-1})/T_s$) smoothed with Exponential Moving Average (EMA).

To enable zero-shot deployment, the specialist library is constructed using a greedy worst-case selection strategy, incentivizing minimally sparse coverage of vehicle dynamics \cite{spotorno2026positioncertifiablestateintegrity, ji2025s}. We employ a standard 6-state Dynamic Single-Track Bicycle Model with Pacejka tire forces \cite{chrosniak2024deep} as the ground-truth generator. Variations between operating regimes are modeled through changes in the parameter set $\Phi$ (e.g., friction, mass, drag), simulating both environmental changes and fleet heterogeneity. The library of frozen neural networks is trained on a hybrid synthetic dataset. This dataset combines uniform state-space sampling with targeted "chirp" trajectories designed to excite high-slip dynamics, ensuring the specialists capture limit-handling behavior often missed by random exploration.

Regarding the Optimal Control Problem, the controller solves a discrete-time optimization over horizon $H=15$ ($T_s = 0.02$ s). The objective minimizes tracking error and control effort: $J = \| \mathbf{p}_H - \mathbf{p}_{\text{ref},H} \|_P^2 + \sum_{k=0}^{H-1} \left( \| \mathbf{p}_{k+1} - \mathbf{p}_{\text{ref},k+1} \|_Q^2 + \| \Delta\mathbf{u}_k \|_R^2 \right)$, subject to the ensemble dynamics $\mathbf{x}_{k+1} = \mathbf{x}_k + T_s \cdot f_{\text{ens}}(\mathbf{x}_k, \mathbf{u}_k, \mathbf{w}_{\text{gov}})$, initial state $\mathbf{x}_0 = \hat{\mathbf{x}}_0$, and standard actuation/rate limits $\mathbf{u} \in [\mathbf{u}_{\min}, \mathbf{u}_{\max}]$ \footnote{Cost matrices are diagonal with $Q_{p}=10$, $R_{\delta}=1.0$, $R_{D}=0.1$. Note that rate constraints apply specifically to steering angle $\delta$ (not throttle $D$) to prevent aggressive steering maneuvers while allowing rapid throttle response.}.  Additionally, the reference trajectory is generated geometrically by projecting positions along the track centerline, assuming a constant longitudinal velocity $v_{\text{ref}} = 1.5$ m/s. This reference ignores friction limits, forcing the NMPC to approximate it as best as possible. The \textit{Governor} module estimates mixing weights  $\mathbf{w}_k$ in runtime. To address the practical constraint that state derivatives ($\dot{\mathbf{x}}$) are rarely measured directly in vehicular systems, we formulate the regression using a first-order forward Euler discretization over a sliding window of recent state-input measurements $\mathcal{H}_k$, where each element $h_k = (\mathbf{x}_{k}, \mathbf{x}_{k-1}, \mathbf{u}_{k-1})$. The \textit{Governor} minimizes $\min_{\mathbf{w}_k} \sum_{h_k \in \mathcal{H}_k} \left\| \mathbf{x}_k - \mathbf{x}_{k-1} - dt\sum_{i=1}^{N} w_{k}^i \Psi^i (\mathbf{x}_{k-1}, \mathbf{u}_{k-1}) \right\|^2_2 $, subject to $\sum_{i=1}^{N} w_k^i = 1, w_k^i \geq 0$. The simplex constraints ensure that the resulting dynamics remain within the convex hull of the specialists. This is an implicit adaptation method, where system behavior is adjusted through coefficient blending over a fixed library of models, enabling adaptation to compound regime changes represented within the specialist ensemble without explicit estimation of physical parameters.

Before injection into the NMPC, the estimated weights $\mathbf{w}_k$ are EMA-smoothed with decay $\alpha=0.1$ to suppress high-frequency oscillations that could destabilize the closed loop during transients. Regression uses a sliding window of $W=20$ recent state-input pairs, balancing robust estimation with responsiveness to abrupt regime shifts. These parameters were tuned empirically; formal sensitivity analysis is left as future work to optimize the noise-rejection versus adaptation-delay trade-off. The selected weights $\mathbf{w}^*_k$ are treated as fixed parameters within one NMPC execution cycle. Under this assumption, the effective model Jacobian is given by a convex combination of the specialists’ Jacobians, $    \mathbf{J}_{\text{eff}}(\mathbf{x}, \mathbf{u}) = \sum_{i=1}^{N} w^*_i \frac{\partial \Psi_i}{\partial \mathbf{x}}(\mathbf{x}, \mathbf{u}),$ allowing analytic gradients to be supplied without differentiating through the weight selection mechanism. We distinguish here between \textit{structural transparency} (traversable symbolic graphs) and \textit{semantic interpretability}. While the solver has full access to the symbolic Jacobian (structural), MLP weights remain semantically opaque, even if interpretability is improved by Physics-Informed Machine Learning techniques. Nevertheless, we focus on quantifying the structural graph traversal cost. To maintain this structural auditability, we implement the ensemble using fully symbolic representations in CasADi~SX rather than opaque compiled code. Consequently, the evaluation of constraint Jacobians and Lagrangian Hessians requires repeated traversal of large symbolic computational graphs induced by the neural network specialists. 

\textbf{Specialist Training.} We train eight frozen Physics-Informed Neural Network (PINN) models, empirically selected to balance coverage of the physical parameter manifold against memory and computational cost. Each specialist approximates continuous-time dynamics $\dot{\mathbf{x}}=\Psi_i(\mathbf{x},\mathbf{u})$ for a distinct operating regime. All specialists are implemented as MLPs (3 hidden layers, 64 neurons, $\tanh$ activations) with Z-score normalized state–control inputs. Training minimizes a composite loss $L_{\text{total}} = L_{\text{data}} + \lambda L_{\text{phy}}, \qquad \lambda=0.1,$ where $L_{\text{phy}}$ enforces kinematic consistency by evaluating constraints on the raw physical state and normalizing them to the network output distribution. Optimization follows a two-phase hybrid protocol: Adam global search (lr $=10^{-3}$, early stopping on validation) followed by L-BFGS refinement initialized from the best Adam weights. This procedure reduces specialist RMSE from $\mathcal{O}(5\times10^{-2})$ (Adam-only) to $\approx 4\times10^{-6}$, limiting epistemic error accumulation over the NMPC horizon. 

We conduct the evaluation in two phases \footnote{All closed-loop and computational experiments are evaluated over 100 independent random seeds. Reported metrics are means with 95\% bootstrap confidence intervals (1000 resamples). Deterministic baselines exhibit zero variance by construction; when relevant, pairwise significance ($p<0.05$) is reported alongside effect sizes. All experiments are conducted in simulation on an Intel Core i7-14700 CPU (2.10 GHz) with 32 GB RAM using CasADi~3.6.x and IPOPT~3.14.x with the MA57 linear solver. The complete implementation repository of this PoC experiments, containing all details for reproducibility will be made available.}:

\textbf{Phase I Computational Benchmarks:} This phase evaluates the solver wall time and adaptation latency for neural versus analytic dynamics models. It utilizes the simulated 1:10 scale touring car (F1TENTH) and 1:43-scale RC car (ORCA) models \cite{JainBayesRace2020} with a horizon $N=15$ and sampling time $dt=0.02$ s. Metrics include a decomposition of NLP evaluation versus linear solve time, and a comparison of controller update latencies following an abrupt friction change ($\mu:1.0\rightarrow1.25$) across three model classes: QP-based neural ensemble reweighting, naive/JIT physics recompilation, and optimized parametric physics updates.

\textbf{Phase II Functional Plasticity Validation:} This phase evaluates closed-loop performance exclusively on the ETHZ 1:43-scale autonomous racing track \cite{JainBayesRace2020}. The track’s tight geometry amplifies small prediction errors, serving as a stress test to expose how epistemic uncertainty propagates under aggressive control. Acknowledging that the identified solver latency currently precludes real-time hardware deployment, we employ a synchronous simulation to decouple functional correctness from computational overhead. In this mode, virtual time pauses during inference so all controllers operate with identical information latency. This proof-of-concept 
evaluation isolates the architectural contributions from real-time implementation constraints, establishing a functional baseline for future deployment. This phase follows a diagnostic hierarchy: (i) ideal ODE specialists defining a performance ceiling, (ii) noisy ODE specialists (additive Gaussian noise, $\sigma=0.05$, same order as Adam-only RMSE) to isolate aleatoric sensitivity, and (iii) neural PINN specialists representing the performance floor. Controllers are evaluated under (i) a sudden wet-track regime shift ($\mu\rightarrow0.5$), (ii) a compound fleet-heterogeneity shift ($\mu\rightarrow0.5$, $m\rightarrow1.2m_{\text{nom}}$, $C_d\rightarrow1.4C_{d,\text{nom}}$), and (iii) a nominal stability case to verify the \textit{Governor} does not induce spurious adaptation.

\section{Results}
\label{sec:results}

Table \ref{tab:combined_compact} synthesizes the trade-offs between symbolic auditability and speed. While the Neural Ensemble enables plasticity, it incurs a $72\times$--$102\times$ slowdown vs. optimized physics baselines. Detailed profiling reveals this is structural: for F1TENTH, NLP evaluation consumes $\sim$588 ms (\textbf{96\%} of total time), while the linear solver takes only $\sim$9 ms ($p<0.001$). Unlike analytical models, the Ensemble forces the solver to traverse large computational graphs to generate verifiable Jacobians. This creates a "cost to compute zeros": the AD engine \cite{andersson2019casadi} must traverse dense paths to verify sparse contributions (consistent $\sim$3.6\% density across both platforms). This consistency across disparate vehicle scales demonstrates that the transparency penalty is a fixed structural tax of the neural architecture, independent of the vehicle's physical mass.

Despite steady-state overhead, the adaptation mechanism is rapid ($\sim$7.3 ms), contrasting with naive "Just-In-Time" recompilation ($\sim$752 ms latency spike). This overhead suggests the specialist networks are overparameterized for the low-dimensional vehicle dynamics ($v_x, v_y, \omega$), pointing to future opportunities for sensitivity-based pruning.

\begin{table*}[t!]
\centering
\caption{Combined Experimental Results. \textbf{(Top) Steady-state performance:} Neural Ensemble incurs slowdowns but achieves 7.3\,ms adaptation latency, outperforming JIT (752\,ms). \textbf{(Bottom) Post-shift RMSE:} Parentheses denote no-shift metrics; positive mitigation indicates improvement over baseline (Tier 1 Base: $v_x$=0.3124, $v_y$=0.0563, pos=0.0602\,m).}
\label{tab:combined_compact}
\vspace{-8pt}

\resizebox{\textwidth}{!}{%
\begin{tabular}{@{}llccccc@{}}
\toprule
\textbf{Testbed} & \textbf{Model} & \textbf{Build Time} (ms) & \textbf{Solve Time} (ms) [95\% CI] & \textbf{Slowdown} & \textbf{Jac. Density} & \textbf{Adaptation Latency} (ms) \\ \midrule
\textbf{F1TENTH} (1:10 Scale) & Physics (Parametric) & 265 & 7.38 [7.20--7.56] & 1.0$\times$ & 3.54\% & $\sim$0.01 (Explicit) / $\sim$752 (JIT) \\
 & Neural Ensemble & 10{,}521 & 535.00 [518--552] & 72.5$\times$ & 3.64\% & \textbf{7.30 [7.1--7.5]} \\ \midrule
 \textbf{ORCA} (RC Car) & Physics (Parametric) & 285 & 5.35 [4.69--5.82] & 1.0$\times$ & 3.54\% & --- \\
& Neural Ensemble & 10{,}850 & 545.00 [530-572] & 101.8$\times$ & 3.64\% & --- \\
 \bottomrule
\end{tabular}%
}

\vspace{2pt}

\resizebox{\textwidth}{!}{
\begin{tabular}{l|ccc|ccc|ccc}
\toprule
Tier & \multicolumn{3}{c}{$v_x$ (m/s)} & \multicolumn{3}{c}{$v_y$ (m/s)} & \multicolumn{3}{c}{position (m)} \\
 & Post B & Post A & Mit. (\%) & Post B & Post A & Mit. (\%) & Post B & Post A & Mit. (\%) \\
\midrule
\multicolumn{10}{c}{\textbf{Friction Only (Severe)}} \\
1 & 0.2374 (0.1617) & 0.2371 (0.1702) & +0.4 (-5.6) & 0.3579 (0.0646) & 0.2566 (0.0593) & +33.6 (+63.9) & 0.1214 (0.0389) & 0.0617 (0.0333) & \textbf{+97.5 (+26.3)} \\
2 & 0.2286 (0.1593) & 0.2352 (0.1742) & -7.9 (-9.7) & 0.3427 (0.0643) & 0.2609 (0.0593) & +28.6 (+62.5) & 0.1134 (0.0390) & 0.0653 (0.0338) & \textbf{+90.4 (+24.5)} \\
3b & 0.2476 (0.2128) & 0.2547 (0.2137) & -11.0 (-0.9) & 0.3102 (0.0634) & 0.2942 (0.0705) & +6.3 (-100.0) & 0.0927 (0.0407) & 0.0762 (0.0386) & \textbf{+50.8 (+10.8)} \\
3a & 1.62 (1.40) & 1.09 (1.05) & +40.5 (+32.2) & 0.78 (0.40) & 0.23 (0.15) & +76.0 (+72.7) & 2.85 (2.00) & 1.12 (0.95) & \textbf{+62.0 (+54.1)} \\
\midrule
\multicolumn{10}{c}{\textbf{All Parameters (Severe)}} \\
1 & 0.3540 (0.1617) & 0.2052 (0.1702) & +357.7 (-5.6) & 0.5598 (0.0646) & 0.5459 (0.0593) & +2.8 (+63.9) & 0.5777 (0.0389) & 0.2882 (0.0333) & \textbf{+55.9 (+26.3)} \\
2 & 0.3512 (0.1597) & 0.2221 (0.1720) & +332.7 (-8.1) & 0.5596 (0.0657) & 0.5385 (0.0595) & +4.2 (+66.0) & 0.5769 (0.0394) & 0.3197 (0.0332) & \textbf{+49.8 (+29.8)} \\
3b & 0.2887 (0.2128) & 0.2569 (0.2137) & +134.2 (-0.9) & 0.5132 (0.0634) & 0.5080 (0.0705) & +1.1 (-100.0) & 0.4870 (0.0407) & 0.3956 (0.0386) & \textbf{+21.4 (+10.8)} \\
3a & 1.54 (1.35) & 0.93 (0.90) & +49.7 (+43.4) & 0.90 (0.45) & 0.28 (0.18) & +73.5 (+68.6) & 2.65 (1.90) & 0.98 (0.85) & \textbf{+64.5 (+57.1)} \\
\bottomrule
\end{tabular}
}
\end{table*}

\begin{figure*}[ht!]
    \centering
    \vspace{-10pt}
    \includegraphics[width=\textwidth]{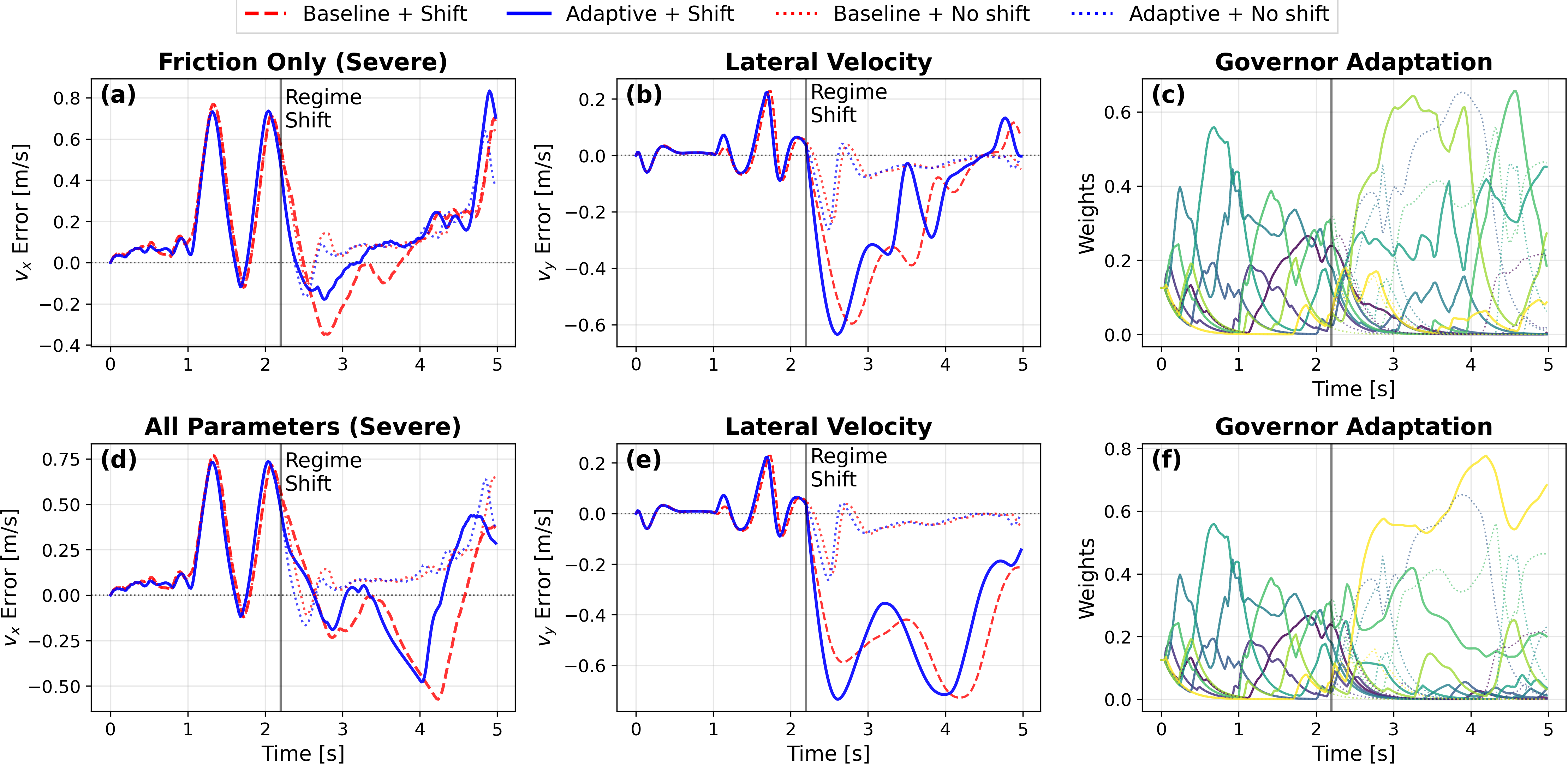}
    \vspace{-20pt}
    \caption{Well-Trained PINN performance, Hybrid training ($\text{RMSE}\approx4\times10^{-6}$) closes gap to ODE ceiling. Position error $<0.1$ m, recovery in 0.8 s.}
    \label{fig:pinn_new}
\end{figure*}

Furthermore, we validate that the proposed architecture solves the vehicular plasticity problem, handling abrupt regime shifts and fleet heterogeneity, provided that the computational bottleneck identified earlier is decoupled via synchronous simulation. Performance is evaluated across three diagnostic tiers, summarized in Table~\ref{tab:combined_compact}. Performance is standardized to Tier 1 pre-shift baseline  ($v_x=0.3124$\,m/s, $v_y=0.0563$\,m/s, position=$0.0602$\,m) with \textit{Degradation} $\delta = 100 \times (\text{post} - \text{pre}_{\text{base}}) / |\text{pre}_{\text{base}}|$ and \textit{Mitigation} $\mu = 100 \times (\delta_{\text{base}} - \delta_{\text{adapt}}) / |\delta_{\text{base}}|.$

Beginning with the theoretical validation in the idealized ODE setting (Tier~1), the \textit{Governor} is expected to recover near-ceiling performance. This is confirmed in Table~\ref{tab:combined_compact}: under \textit{Friction Only} shifts, the adaptive controller contains position degradation to $+12.9\%$, achieving 97.5\% mitigation relative to the baseline. Introducing aleatoric uncertainty via sensor noise in Tier~2 ($\sigma=0.05$) demonstrates that despite high-frequency jitter in the residuals, the regression logic remains stable. Position mitigation remains high, and weight trajectories remain smooth, confirming that aleatoric uncertainty does not destabilize the \textit{Governor’s} estimation mechanism, consistent with the EMA smoothing. When instantiated with white-box neural specialists (Tier~3), performance becomes strictly governed by epistemic uncertainty. In the case of naively trained PINNs (Tier~3a) using Adam-only training, absolute prediction errors are unsafe (meter-scale position drift). Nevertheless, adaptation functions as a safety layer: as shown in Table~\ref{tab:combined_compact}, the \textit{Governor} mitigates the dominant failure modes, reducing lateral velocity error and preventing spin-out during high-curvature maneuvers. This arises from specialist reweighting that corrects systematic lateral slip underestimation and partially compensates biased longitudinal drag, though residual bias induces oscillatory weights.

However, with the hybrid training protocol in Tier~3b, epistemic uncertainty is minimized and the architecture closes the gap to the ODE ceiling. As illustrated in Figure~\ref{fig:pinn_new}, position error remains below $0.1$~m throughout operation, matching the ideal ODE case. Quantitatively, this corresponds to a $35\times$ reduction in absolute tracking error relative to the unsafe Tier~3a baseline. In the \textit{All Parameters} heterogeneity test, including simultaneous shifts in friction, mass, and drag, the \textit{Governor} still recovers a significant fraction of performance. While the lower mitigation relative to the friction-only case reflects the limits of library coverage for simultaneous parameter excursions, the system still yields substantial robustness absent in non-adaptive baselines. Null-hypothesis validation across all tiers confirms that no-shift conditions exhibit bounded or negative degradation, with position error often improving over time due to a mild regularization effect from convex blending. While the \textit{Governor} continuously makes minor weight adjustments in response to tracking-induced residuals, significant changes occur primarily during genuine regime shifts (Figure~\ref{fig:pinn_new}), confirming stable and non-spurious adaptation, a key prerequisite for safe deployment. Thus, the transition from naive Tier~3a to hybrid Tier~3b closes the safety gap by more than an order of magnitude, recovering the performance ceiling of the ideal ODE specialists. This functional gain justifies the computational overhead identified: although symbolic graph traversal incurs a $72$--$102\times$ slowdown depending on the baseline physics complexity, it enables a verified, white-box controller capable of autonomous adaptation with sub-lane precision. The functional validation confirms that the architecture successfully solves the plasticity problem, motivating future work to optimize the computational implementation to make this trade-off viable in real-time.

\section{Discussion and Conclusion}
\label{sec:discussion}

This work quantifies the efficiency cost of prioritizing symbolic auditability over execution speed in white-box adaptive control. Results reveal that strict synchronous verifiability currently prohibits real-time deployment, yet the architecture functionally solves vehicular plasticity. The observed 72--102$\times$ slowdown relative to parametric physics establishes the structural price, and empirical baseline, of maintaining full runtime introspection. Unlike latency-optimized approaches \cite{2502.02310,2304.12263}, our work isolates the structural cost of symbolic neural ensemble auditability. While code-generation frameworks such as \textit{acados} or L4CasADi compile models into opaque binaries, our approach forces the solver to traverse the explicit symbolic computation graph of the neural ensemble at every iteration. This ensures that every Jacobian entry is traceable to a specific specialist contribution, providing a level of introspection deeper than standard tool qualification. Although, it currently introduces a ``cost to compute zeros'' where the AD engine spends the majority of solver time traversing sparse graph paths.

Despite this computational overhead, the functional validation confirms that the resulting plasticity provides critical robustness where parametric models fail. While optimized physics models excel with pre-defined parameters, they are brittle to unmodeled heterogeneity. Our implicit adaptation handles compound regime shifts in friction, mass, and drag, reducing tracking error compared to non-adaptive baselines. Crucially, this plasticity is achieved without the instability risks of online backpropagation, as the \textit{Governor} operates solely by blending pre-verified, frozen specialists. This confirms that the computational cost buys a safety margin: the ability to handle epistemic uncertainty without leaving the verified convex hull of the model library. Furthermore, the observed performance improvement in no-shift scenarios suggests that this convex blending acts as an ensemble regularizer, smoothing out minor model mismatches better than a single parametric model.

Consequently, these findings do not suggest a dead end, but rather a roadmap for making white-box adaptation real-time viable. The consistent low Jacobian density ($\sim$3.6\%) across scales indicates that sensitivity-based pruning could compress the symbolic graph by an order of magnitude without functional loss. Furthermore, the significant decoupling of adaptation latency ($\sim$7.3\,ms) from the heavier solver latency implies that an asynchronous architecture could reconcile the conflict between plasticity and speed, though this may require trading strict synchronous auditability for performance. Future work must also address the theoretical and semantic limitations identified in this diagnostic study. While our empirical results validate the geometric constraints of the convex hull, formal certification will require deriving dwell-time conditions or Lyapunov bounds for the \textit{Governor's} switching logic to guarantee closed-loop stability. Finally, to bridge the gap between structural transparency (traversable graphs) and true semantic interpretability, future iterations should replace opaque MLPs with Kolmogorov-Arnold Networks (KANs), aligning the physical meaning of the specialists with the inspectability of the solver. Thus, this study serves as a proof-of-concept and a baseline: while synchronous white-box ensembles currently hit an efficiency wall, they offer a verifiable, transparent path to solving the plasticity-stability dilemma.

\bibliographystyle{IEEEtran}
\bibliography{ref}

\end{document}